\documentclass[sigconf]{acmart}
\AtBeginDocument{%
  }

\usepackage{colortbl}
\usepackage{enumitem}
\usepackage{hyperref}
\usepackage{url}
\usepackage{wrapfig}
\usepackage{booktabs}

\usepackage{amsfonts}       
\usepackage{amsmath}
\usepackage{graphicx}
\usepackage{amsthm}
\usepackage{subfigure}
\usepackage{pifont}
\usepackage{mathtools}
\usepackage{stmaryrd}
\usepackage[T1]{fontenc}
\usepackage{multirow}
\newcommand{\goodname}{\textsf{Wildflare GuardRail}}
\usepackage[vlined,linesnumbered,ruled,resetcount]{algorithm2e}
\usepackage{wrapfig}
\usepackage{lipsum}  
\usepackage{booktabs}

\definecolor{redx}{RGB}{180,0,0}
\definecolor{greenx}{RGB}{0,180,0}

\newcommand{\redxmark}{\color{redx}\ding{55}}
\newcommand{\greencmark}{\color{greenx}\ding{51}}
\definecolor{redx}{RGB}{180,0,0}
\definecolor{greenx}{RGB}{0,180,0}
\usepackage{flushend}
\usepackage{algpseudocode}  

\newcommand{\grounding}{{\textsf{Grounding}}}
\newcommand{\detection}{{\textsf{Safety Detector}}}
\newcommand{\fixing}{{\textsf{Repairer}}}
\newcommand{\customization}{{\textsf{Customizer}}}
\newtheorem{definition}{Definition}
\newtheorem{example}{Example}
\usepackage{algpseudocode}
\usepackage{amsmath}       

\usepackage{quoting}

\begin{document}

\title{Bridging the Safety Gap: A Guardrail Pipeline for Trustworthy LLM Inferences}

\author{Shanshan Han}
\affiliation{%
  \institution{University of California, Irvine}
  \city{Irvine}
  \state{California}
  \country{USA}
}
\email{shanshan.han@uci.edu}

\author{Salman Avestimehr}
\affiliation{%
  \institution{University of Southern California}
  \city{Los Angeles}
  \state{California}
  \country{USA}}
\email{avestime@usc.edu}

\author{Chaoyang He}
\affiliation{%
  \institution{TensorOpera AI}
  \city{Palo Alto}
  \state{California}
  \country{USA}}
\email{ch@tensoropera.com}

\renewcommand{\shortauthors}{Han et al.}

\begin{abstract}
We present \goodname, a guardrail pipeline designed to enhance the safety and reliability of Large Language Model (LLM) inferences by systematically addressing risks across the entire processing workflow. 
\goodname~integrates several core functional modules, including 
\detection~that identifies unsafe inputs and detects hallucinations in model outputs while generating root-cause explanations,
\grounding~that contextualizes user queries with information retrieved from vector databases, \customization~that adjusts outputs in real time using lightweight, rule-based wrappers, 
and
\fixing~that corrects erroneous LLM outputs using hallucination explanations provided by \detection.
Results show that our unsafe content detection model in \detection~achieves comparable performance with OpenAI API, though trained on a small dataset constructed with several public datasets. 
Meanwhile, the lightweight wrappers can address malicious URLs in model outputs in 1.06s per query with 100\% accuracy without costly model calls. Moreover, the hallucination fixing model demonstrates effectiveness in reducing hallucinations with an accuracy of 80.7\%. 

\end{abstract}

\maketitle

\section{Introduction}

Large Language Models (LLMs) are increasingly deployed in latency-sensitive, high-stakes systems—from network automation to real-time decision support in critical fields such like healthcare~\cite{goyal2024healai,qiu2024llm,yang2024talk2care} and finance~\cite{wu2023bloomberggpt,li2023large}. However, their widespread adoption is hindered by significant safety risks. Malicious inputs can exploit prompt injection vulnerabilities to manipulate outputs~\citep{Liu2023PromptIA,Kumar2023CertifyingLS,Zhu2023PromptBenchTE,Chu2024ComprehensiveAO,Tedeschi2024ALERTAC,Zhao2024WeaktoStrongJO}, while unconstrained responses may propagate hallucinations, biases, nonsensical or factually incorrect knowledge, or security threats like phishing URLs~\citep{Zhang2023SirensSI,zhang2024efficient,Wang2023DecodingTrustAC,Fan2023OnTT,Huang2023ASO,Xu2024HallucinationII}. These issues not only compromise user trust but also pose systemic risks, such as resource misuse in cloud networks or erroneous configurations in software-defined infrastructures~\citep{llm_loops,owasp_llm01,owasp_llm03,owasp_llm04,owasp_llm05}.

Safeguarding LLMs is crucial and can never be overstated. Unsafe inputs can manipulate LLM outputs, reveal sensitive information, bypass system instructions, or execute malicious commands~\citep{Russinovich2024GreatNW,xu2024llm,chang2024play,Liu2023PromptIA,Kumar2023CertifyingLS,Zhu2023PromptBenchTE,Chu2024ComprehensiveAO,Tedeschi2024ALERTAC,Zhao2024WeaktoStrongJO}. 
Problematic outputs can confuse users, perpetuate biases, and undermine users' trust in LLM-based systems, particularly in domains like healthcare and finance, where inaccuracies or biases can have legal or societal repercussions.

Addressing safety issues in LLM inference is complex, as risks can arise at any point during processing user queries. 
While existing work addresses isolated aspects of LLM safety~\cite{kumar2024watch,elesedy2024lora,ma2023adapting,jha2024memeguard}, no unified solution holistically mitigates risks across the entire inference pipeline.
Standalone detection models~\citep{Detoxify,perspective-api,openai-data-paper} operate reactively to flag unsafe content, but they require full retraining or finetuning to adapt to new safety requirements and lack mechanisms to correct errors in the LLM outputs. 
Post-hoc correction methods rewrite problematic content in the LLM outputs but fail to address their root causes, such as hallucinations, that often stem from insufficient, inaccurate, or outdated source information~\citep{Xu2024HallucinationII,Zhang2023SirensSI,Huang2023ASO}. While retrieval-augmented generation (RAG)~\cite{rag,chen2024benchmarking,gao2023retrieval} can mitigate hallucinations by enriching user queries with external contextual knowledge, 
the probabilistic retrieval of knowledge cannot enforce deterministic safety policies, e.g., blocking mandated sociopolitical terms. 
Rule-based post-processing ``wrappers''~\citep{guardrails}, on the other hand, offer agility for time-sensitive updates and excel at syntactic filtering (e.g., regex-based phishing URL detection via APIs like Google Safe Browsing~\citep{google-safe-browsing}), but fail to address semantic risks in the contents generated by LLMs.

\begin{table*}[ht]
\centering
\caption{Comparison of moderation-based harmfulness mitigation approaches}
\label{tab:moderation_comparison}
\begin{tabular}{lccccccc@{}} 
\toprule
\textbf{Feature} & \textbf{Perspective API} & \textbf{Open AI} & \textbf{Nvidia NeMo} & \textbf{GuardRails} & \textbf{Detoxify} & \textbf{Llama Guard} & \textbf{Ours} \\ \midrule
Open-sourced &\redxmark&\redxmark&\greencmark &\greencmark &\greencmark &\greencmark &\greencmark \\
Self-developed model &\greencmark &\greencmark &\redxmark&\redxmark&\greencmark &\greencmark &\greencmark \\
Deployable on edge devices &\greencmark &\greencmark & - & - &\greencmark & \redxmark &\greencmark \\
Zero-shot generalization &\greencmark &\greencmark & - & - &\greencmark &\greencmark &\greencmark \\
Explainable results &\redxmark&\redxmark&\redxmark&\redxmark&\redxmark&\redxmark& \greencmark \\
Flexible workflow &\redxmark&\redxmark&\greencmark &\greencmark &\redxmark&\redxmark&\greencmark \\
\bottomrule
\end{tabular}
\end{table*}

\textit{We argue that, enhancing the overall safety of LLM inference 
demands a comprehensive pipeline that orchestrates heterogeneous functional components, such as ML models, RAG, and light-weighted wrappers.} A well-designed guardrail pipeline not only mitigates safety risks from a global perspective but also enable users to customize their workflows to high flexibility and efficiency.

This paper introduces \goodname, a guardrail pipeline that systematically integrates detection, contextualization, correction, and customization to ensure robust safety and adaptability during LLM inference. \goodname~integrates four components, including
\textit{i}) \detection~that identifies unsafe content (e.g., toxicity, bias, hallucinations) in user inputs and LLM outputs; \textit{ii}) \grounding~that contextualize user queries with vector databases; \textit{iii})
\customization~that leverages lightweight wrappers to edit LLM output according to user needs in a real-time manner; and \textit{iv})
\fixing~that corrects hallucinated content detected in the LLM outputs.
Our contributions are summarized as follows:

\begin{itemize}[leftmargin=16pt, itemsep=1pt]
    \item \textit{Holistic safety pipeline. }We propose \goodname, a comprehensive guardrail pipeline for safeguarding LLM inferences. \goodname~incorporates
detection, contextual grounding, output correction, and user customization into a unified workflow, addressing safety issues in LLM inference as a systemic challenge rather than isolated subproblems.

\item \textit{Specialized fine-tuned models. } We utilized our pretrained LLM, Fox-1~\cite{fox}, as the base model and fine-tuned three lightweight models, including  \textit{i}) a  moderation model that detects unsafe content in user inputs and LLM outputs (\S\ref{sec: unsafe_input_detection}); 
\textit{ii}) an explainable hallucination detection model that detects hallucinations in the LLM outputs while providing explanations for the hallucinations (\S\ref{sec: hallucination_detection_and_reasoning}); and \textit{iii}) a fixing model that corrects problematic LLM outputs based on the explanations for hallucinations (\S\ref{sec:fixing}).
These models are light-weighted and can be deployed on edge devices.

\item\textit{Explainable Hallucination Mitigation. } We address the hallucination issue through a two-stage approach that first detects hallucination and pinpoints hallucination causes
with \detection (\S\ref{sec: hallucination_detection_and_reasoning}), and then utilize \fixing~to correct the problematic content based on the reason of hallucination (\S\ref{sec:fixing}).

\item\textit{Effective grounding. } We propose two indexing methods, including \textit{Whole Knowledge Index} and \textit{Key Information
Index}, to assist retrieving knowledge from vector data storage  (\S\ref{sec:grounding}).

\item\textit{Flexible User-defined safety protocols. } \goodname~allows users to define protocols for customizing LLM outputs with
\customization, which is flexible to adapt to evolving user needs while
providing real-time solutions for addressing safety issues in LLM
deployments, without pretraining or fine-tuning models to address
emerging safety challenges (\S\ref{sec:customization}).

\end{itemize}

\section{Related Work}

Moderation-based harmfulness mitigation approaches leverage rule-based methods, ML classifiers, and human interfaces to monitor, evaluate, and manage the outputs produced by LLMs to ensure the outputs generated by LLMs are safe, appropriate, and free from harmful content~\citep{openai-data-paper,nemo,perspective-api,Detoxify,guardrails}.
We compare our approaches with the existing approaches in Table~\ref{tab:moderation_comparison}.

\underline{\textit{Close-sourced solutions. }} 
OpenAI Moderation API~\citep{openai-data-paper} and Perspective API~\citep{perspective-api} utilize ML classifiers to detect undesired contents. These approaches provide scores for pre-defined categories of harmful content, such as toxicity, identity attacks, insults, threats, etc. These tools are widely used in content moderation to filter out harmful content and has been incorporated into various online platforms to protect user interactions~\citep{perspective-api-case-studies}. However, they are less adaptable to emerging safety risks as they are not open-sourced and cannot be finetuned.

\underline{\textit{Opensourced solutions. }} 
LlamaGuard~\citep{inan2023llamaguard} leverages the zero-shot and few-shot abilities of the Llama2-7B architecture~\citep{touvron2023llama} and can adapt to different taxonomies and sets of guidelines for different applications and users. Despite its adaptability, LlamaGuard's reliability depends on the LLM's understanding of the categories and the model's predictive accuracy. However, deploying LlamaGuard on edge devices is challenging due to its large number of parameters, which typically exceed the computing resources available on edge devices.
Detoxify~\citep{Detoxify} offers open-source models designed to detect toxic comments. These models, based on BERT~\citep{devlin2018bert} and RoBERTac~\citep{liu2019roberta} architectures, are trained on the Jigsaw datasets~\citep{jigsaw-unintended-bias-in-toxicity-classification,jigsaw-toxic-comment-classification,jigsaw-multilingual}. Detoxify provides pre-trained models that can be easily integrated into other systems to identify toxic content. Also, the models are able to recognize subtle nuances in language that might indicate harmful content, making them effective for moderation.

\underline{\textit{Customizable solutions. }}
Guardrails~\citep{guardrails} and Nvidia NeMo~\citep{nemo} employ customizable workflows to enhance safety in LLM inference. 
Guardrails~\citep{guardrails} define flexible components, called ``rails'', to enable users to add wrappers at any stage of inference, which enables  users to add structure, type, and quality guarantees to LLMs outputs. Such rails can be code-based or using ML models. However, it does not have self-developed model and miss a unified solution for general cases. 
Nvidia NeMo Guardrails~\citep{nemo} functions as an intermediary layer that enhances the control and safety of LLMs. 
This framework includes pre-implemented moderation dedicated to fact-checking, hallucination prevention, and content moderation, which offers a robust solution for enhancing LLM safety.

\begin{figure*}
  \centering
  \includegraphics[width=0.92\textwidth]{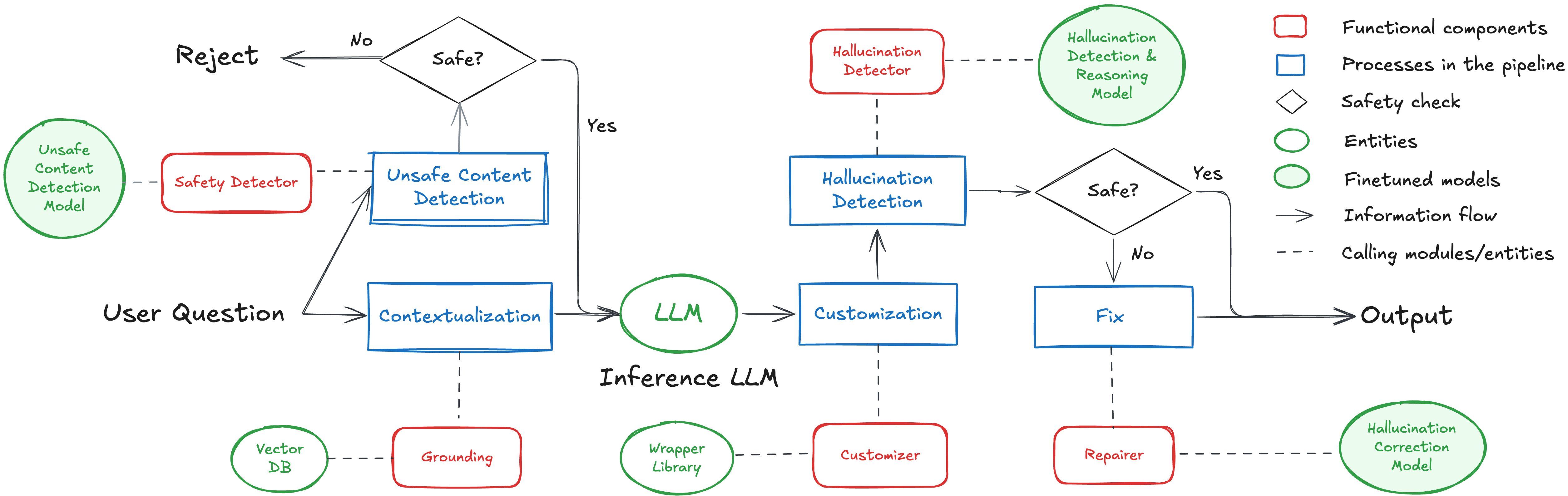}
  \caption{Overview.}
  \label{fig: system_overview}
\end{figure*}

\section{\goodname~Overview}
\goodname~enhances safety of LLM inputs and outputs while improving their quality. Specifically, it achieves two goals, 1) all user inputs are safe, contextually grounded, and effectively processed, such that the inputs to the LLMs are of high-quality and informative; and 2) the output generated by the LLMs are evaluated and enhanced, such that the outputs passed to users can be both relevant and of high quality. 
The pipeline can be partitioned into two parts, including 
1) processing before LLM inference that enhances user queries, and 2) processing after LLM inference that detects undesired content and handle them properly. We overview our pipeline in Figure~\ref{fig: system_overview}.

\noindent\underline{\textit{Pre-inference processing. }}
Before sending user queries to LLMs, \goodname~detects if there are any safety issues in the queries with \detection~and ground the queries with context knowledge with \grounding. 
\detection~monitors user inputs to identify and reject queries that might be unsafe. The monitoring includes typical safety checks, including toxicity, stereotypes, threats, obscenities, prompt injection attacks, etc. Any form of unsafe content will lead to the queries being rejected. 
Inputs that pass this initial safety check are grounded with context with \grounding, where the user query is contextualized and enhanced with relevant knowledge retrieved from the vector data storage. By equipping the query with some context knowledge, the LLM can do inference with enriched information, thus can reduce hallucinations when generating responses. The details of \detection~ and \grounding~will be introduced in \S\ref{sec:safety_detector} and \S\ref{sec:grounding}, respectively.

\noindent\underline{\textit{Post-inference processing. }}
Upon LLM finishing inference, \detection~detects safety issues in the LLM outputs, specifically, hallucinations. This is because LLM applications typically leverages well-developed LLMs or APIs, such as LLaMA~\citep{touvron2023llama} and ChatGPT API~\citep{openai-data-paper}, which are generally safe and less likely to generate toxic or other unsafe content, while hallucinations occur frequently. \detection~identifies hallucinations and provides reasons for the hallucinations, such that \goodname~can utilize the reasoning for later refinement of the LLM outputs. To achieve goal, \goodname~employs a text generation model to generate explainable results, and adjusts the loss function during training to ensure the model to produce classification results. 
After \detection~finishes detection, \fixing~fixes the problematic content or aligns the outputs with some rule-based wrappers to meet user expectations. 
If the outputs are difficult to fix, e.g., hallucinated responses, 
\fixing~will call a fixing model to fix the answers. Details about \fixing~can be found in \S\ref{sec:fixing}.

\vspace{-1em}

\section{\goodname~\detection}\label{sec:safety_detector}

\detection~addresses unsafe inputs and inappropriate LLM responses to ensure that both the user queries provided to the models and the LLM outputs are safe and free from misinformation. 
\subsection{Unsafe Input Detection}\label{sec: unsafe_input_detection}

We developed a model to detect unsafe contents in user queries before they are processed by LLMs for inference.
While existing approaches categorize unsafe content into various types (e.g.,  toxicity, prompt injection, stereotypes, harassment, threats, identity attacks, and violence)~\citep{openai-data-paper,Wang2023DecodingTrustAC,Detoxify}, our method employs a unified, binary classification model finetuned based on our opensourced LLM~\citep{fox}, classifying content as  safe or unsafe.

This strategy offers several key advantages, as follows: \textit{i}) By fine-tuning our base model, which has been trained on vast amounts of data, the classification model can leverage pre-existing knowledge relevant to safety detection.
\textit{ii}) A binary classification of ``safe'' and ``unsafe'' is both efficient and sufficient for LLM services, as any unsafe query should be rejected, regardless of the specific risk.
\textit{iii}) This approach avoids the complexities and potential inaccuracies of categorizing overlapping or ambiguous types of unsafe content in some publicly available datasets. For example, toxicity toward minority groups could also be classified as bias, but current datasets may inadequately capture such nuances.
\textit{iv}) Using straightforward code logic, we can transform public datasets for safety detection into clear safe/unsafe labels, minimizing ambiguity and ensuring high-quality training data.

The biggest challenge in training such model is the discrepancy between the training data and real-world user query distributions, where using traditional datasets alone can result in poor performance due to their divergence from actual user queries~\citep{openai-data-paper}.
To mitigate these issues, we integrated data of various domains and contexts to better simulate the variety of unsafe queries that users might submit.
We crafted a training dataset
by combining samples randomly selected from 15 public datasets, as will be introduced in Table~\ref{tab:exp_datasets} in \S\ref{sec: exp}. 
Such a dataset captures  diverse contents in user inputs in practice, thus can be more representative on potential real-world inputs.

\subsection{Hallucination Detection and Reasoning}\label{sec: hallucination_detection_and_reasoning}

\begin{algorithm}[!t]
\textbf{Inputs:} 
$\mathcal{D}$: a training dataset that contains ``context'', ``inputs'', ``llm\_answer'', and ``labels'' for hallucination; 
\textit{\text{prompt}\_\text{template}}: for formulating the hallucination detection data, see Figure~\ref{fig: training_data_example}; \textit{\text{GPT}\_\textit{reasoning}\_\text{template}}: for generating prompts for GPT API, see Figure~\ref{fig: training_data_example}.

\textbf{Outputs:} $\mathcal{D}_t$: the training dataset.

\nl{\bf Function $\boldsymbol{\mathit{process\_data}(\mathcal{D})}$} \nllabel{ln:function_process_data}
\Begin{

\nl $\mathcal{D}_t\leftarrow \phi$

\nl \For{$d \in $ $\mathcal{D}$}{

\nl \eIf{$\mathit{is}\_\mathit{hallucination}$($d$)}{

\nl $\mathit{halu}\_\mathit{reason}\leftarrow\mathit{GPT}\_\mathit{API}$(\textit{GPT}\_\textit{reasoning}\_\textit{template}($d$[``$\mathit{question}$''], $d$[``$\mathit{context}$''], $d$[``$\mathit{llm}\_\mathit{answer}$'']))

\nl $\mathit{response}\leftarrow$ ``Yes, '' + $\mathit{halu}\_\mathit{reason}$

\nl $d^\prime\leftarrow$\textit{prompt}\_\textit{template}($d$[``$\mathit{question}$''], $d$[``$\mathit{context}$''], $d$[``$\mathit{llm}\_\mathit{answer}$''], $\mathit{response}$)

}{

\nl $d^\prime\leftarrow$\textit{prompt}\_\textit{template}($d$[``$\mathit{question}$''], $d$[``$\mathit{context}$''], $d$[``$\mathit{llm}\_\mathit{answer}$''], ``$\mathit{No.}$'')

}

\nl $\mathcal{D}_t$.$\mathit{add}(d^\prime)$

}

\nl \textbf{return $\mathcal{D}_t$}
}

\caption{Hallucination detection training data processing.}
\label{alg:hallucination_data_processing}
\end{algorithm}

Hallucinations occur when the LLM generates responses that is inaccurate, fabricated, or irrelevant~\citep{filippova2020controlled, maynez2020faithfulness,huang2023survey,rawte2023survey}.
Despite appearing coherent and plausible, hallucinated LLM responses are unreliable, often containing fabricated, misleading information that is  
divergent from the user input, thus fail to meet users' expectations and severely undermine the trustworthiness and utility of the LLM applications.
While grounding can mitigate hallucinations by contextualizing user inputs and enriching the informativeness of user queries, it cannot eliminate hallucinations entirely. 
This is because hallucinations stem from nearly every aspects of LLM training and inference, such as low-quality training data~\citep{lin2021truthfulqa,kang2023impact} and 
randomness of sampling strategies~\citep{chuang2023dola}, and moreover, the very nature probabilistic properties of LLMs. 

Effectively handling hallucinations in LLM responses is both crucial and challenging for producing high-quality LLM responses. 
Existing works that detect presence of hallucinations are insufficient~\citep{manakul2023selfcheckgpt,liu2021token}. To provide high-quality responses to users,  we should handle the detected hallucinations properly, i.e., obtaining the explanations for the hallucinations in the LLM responses and further, fixing the hallucinated responses if possible.

To this end, we propose utilizing our own LLM, Fox-1, as base model~\citep{fox} to finetune a 
hallucination detection model  for detecting hallucinated content and providing explanations, and further, facilitating the subsequent \fixing~in \S\ref{sec:fixing}. The design of the model has the following advantages: \textit{i}) \textit{classification}: it identifies the presence of hallucinations in the LLM output; and \textit{ii}) \textit{reasoning}: it generates explanations for the hallucinated contents, offering insights for the subsequent correction in \fixing; \textit{iii}) \textit{simultaneous classification and reasoning}: it process \textit{i}) and \textit{ii}) at the same time, which saves computation cost and improves efficiency; and \textit{iv}) \textit{vast pre-training data}: it leverages pre-existing knowledge
on hallucination in the base model, which may potentially benefit hallucination detection and reasoning.

\begin{figure*}
  \centering
  \includegraphics[width=\textwidth]{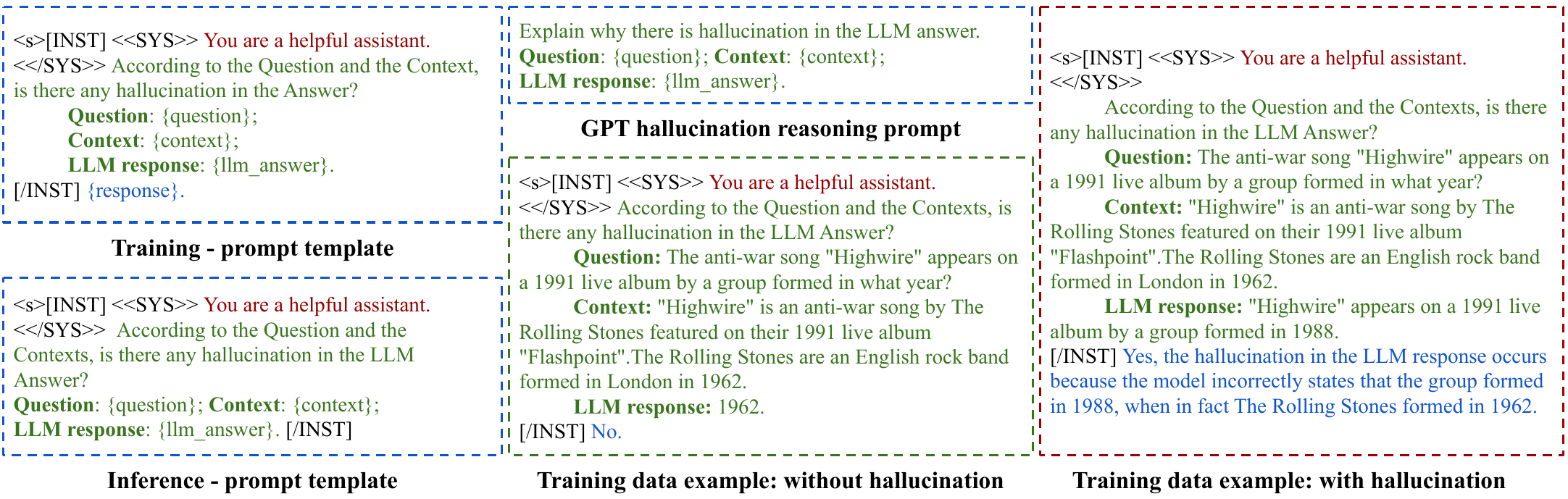}
  \caption{Prompt templates and sample training data for hallucination detection and reasoning.}
  \label{fig: training_data_example}
\end{figure*}

\textbf{Training. }
We feed our base model with hallucination dataset to train a model for both detecting and reasoning for the hallucination. 
However, public available datasets for hallucinated LLM responses are mainly classification datasets with texts and labels, e.g., HaluEval~\citep{li2023halueval}. To address this, we utilize the GPT4 API~\citep{openai-data-paper} to generate explanations for hallucinated contents, and
define a prompt template
to create structured prompts based on the classification data to make it suitable for classification and reasoning simultaneously. 
We demonstrate the prompt templates and sample training data in Figure~\ref{fig: training_data_example}, and summarize data processing in Algorithm~\ref{alg:hallucination_data_processing}.

\begin{algorithm}[!t]
\textbf{Inputs:}
$\mathcal{M}$: hallucination detection model; 
$\mathit{tokenizer}$: tokenizer for $\mathcal{M}$; $\mathit{q}$: a query submitted by users;
$\mathit{context}$: the context to answer the question; retrieved from vector data storage;
$a$: the answer returned by an LLM for the question; \textit{inference}\_\textit{prompt}\_\textit{template}: see Figure~\ref{fig: training_data_example}.

\nl{\bf Function $\boldsymbol{\mathit{inference}(\mathcal{M}, \mathit{q}, \mathit{context}, a, k)}$} \nllabel{ln:inference}
\Begin{

\nl $\mathit{prompt}\leftarrow$\textit{inference}\_\textit{prompt}\_\textit{template}($\mathit{q}$, $\mathit{context}$, $a$)

\nl $\mathit{tokenized}\_\mathit{prompt}\leftarrow \mathit{tokenizer}$($\mathit{prompt}$)

\nl $\mathit{halu}\_\mathit{res}\leftarrow$ 
$\mathcal{M}$.$\mathit{generate}$($\mathit{tokenized}\_\mathit{prompt}$)

\nl $\mathit{first}\_\mathit{word}\_\mathit{logits}\leftarrow \mathit{halu}\_\mathit{res}$.$\mathit{logits}$[0], 

\nl$\mathit{results}\leftarrow\mathit{softmax}$($\mathit{first}\_\mathit{word}\_\mathit{logits}$)

\nl $\mathit{top\_k\_probs}\leftarrow \mathit{top}$($\mathit{results}$, $k$)

\nl $P_{\mathit{halu}}(a)\leftarrow \mathit{compute}\_\mathit{halu}\_\mathit{prob}(\mathit{top}\_k\_\mathit{probs})$ 


\nl \lIf{$P(a)\geq 0.5$}{\textbf{return} True}

\nl{\textbf{return} False}
}

\caption{Hallucination detection model inference.}
\label{alg:inference}
\end{algorithm}

\textbf{Inference. }
We expect the LLM to directly output results whether the LLM response contains hallucinations, \textit{i}.\textit{e}., the first token of outputs to be ``Yes'' or ``No'' as detection results, according to the formatted data sample in Figure~\ref{fig: training_data_example}. However, the first token of the LLM response is probabilistic due to the self-autoregressive nature of decoder-based text generation LLMs. 
To obtain desired outputs, we formulate the text-generation outputs by utilizing the top-$k$ first tokens (and their possibilities) of the outputs to generate classification results. By default, $k$ is 10. 

\begin{definition}[Probability of hallucination]
Let $a$ be an LLM answer, let $\{t_1, ..., t_{k}\}$ be the top-k potential first token, and let $\{p_1, ..., p_{k}\}$  be their top-k probabilities. Let $T$ be a tokenization function, and let $T(\text{"Yes"})$ and $T(\text{"No"})$ be the tokens corresponding to ``Yes'' and ``No'', respectively. The probability of hallucination in $a$ is
$P_\mathit{halu}(a) = \frac{\sum_{i=1}^{k} P(t_i | t_i\in T(\text{"Yes"}))}{\sum_{i=1}^{k} P(t_i | t_i\in T(\text{"Yes"})) + \sum_{i=1}^{k} P(t_i | t_i\in T(\text{"No"}))}
$   
\end{definition}\label{def:halu_prob}

Detection results with $P_{\mathit{halu}}(*)\geq0.5$ indicate the content is classified as ``hallucinated''; otherwise, the content is ``safe''.  The detailed procedure of  inference is described in Algorithm~\ref{alg:inference}.

\section{\goodname~\grounding}\label{sec:grounding}

\goodname~\grounding~enhances the contextual richness and informativeness of user queries by leveraging external knowledge in vector database. Thus, LLMs can utilize such contextual knowledge to generate high-quality outputs, particularly by grounding user queries before they are passed to the LLMs for inference.

To support similarity search over the knowledge data, \goodname~creates vector indexes by vectorizing plaintext knowledge.
\goodname~employs two primary methods for indexing: \textit{i}) \textit{{Whole Knowledge Index}} that creates indexes based on each entire data entry in the datasets; and \textit{ii}) \textit{{Key Information Index}} that indexes only the key information in each data entry, i.e., questions in QA datasets. 
Whole Knowledge Index reflects the data distribution and ensurers that the indexed data captures the contextual variety and complexity found in real-world queries, while Key Information Index 
focuses on the core information of each data entry, thus facilitates efficient retrieval of relevant data. 
We evaluate the effectiveness of indexes with \textit{callback}, i.e., the probability of successfully retrieving the original records from a dataset using Top-$k$ queries. 
We experimentally evaluate the indexing methods in \S\ref{sec: exp}.

\begin{definition}[Callback]
    Let $D_v$ be a vector data storage that contains $n$ records, let $Q$ be a plaintext user query set, and let $I(Q)$ be the vector index created based on $Q$. For each query $q\in Q$, let $I_q$ be the vector index created based on $q$, and let $D_v(I_q)$ denote the set of Top-$k$ records returned by querying $D_v$ with $I(q)$, and let $r_q$ denote the most relevant record of $q$ in $D_v$. 
    The callback for Top-$k$ queries on the query set $Q$ is defined as:
$$C_k(Q) = \frac{1}{|Q|} \sum_{q \in Q} [r_q \in D_v(I_q)]$$
where $[\cdot]$ is Iverson Bracket Notation~\citep{iverson1962programming}, equal to 1 if the condition inside is true, and 0 otherwise.
\end{definition}

To ensure effective and informative grounding, 
the distribution of the index should closely align with query patterns, i.e., query distributions. 
By grounding user queries with knowledge retrieved with a proper index, 
the LLMs can generate contextually appropriate responses, and further, reduce hallucinations and improve the quality of the responses.

\begin{figure*}
  \centering
  \includegraphics[width=0.86\textwidth]{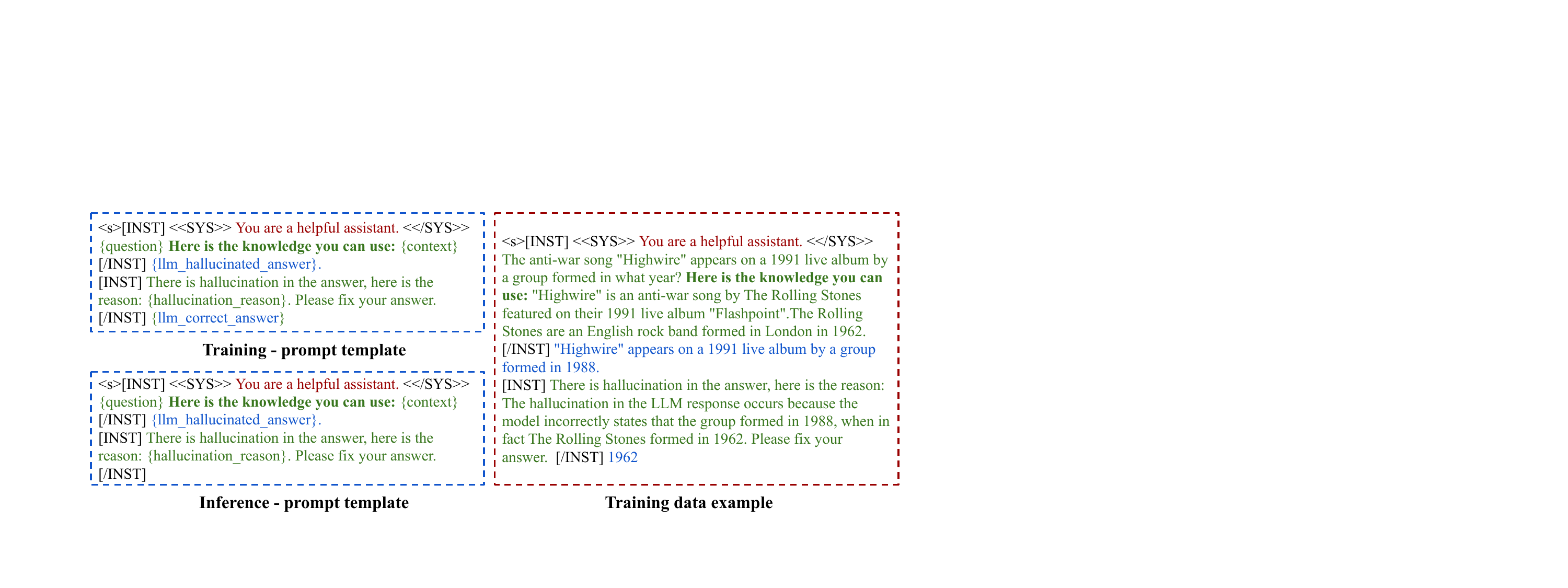}
  \caption{Prompt templates and sample training data for \fixing.}
  \label{fig: training_data_example_fixing}
\end{figure*}

\section{\goodname~\customization}\label{sec:customization}

\goodname~\customization~utilizes lightweight wrappers to flexibly edit or customize LLM outputs to fix some small errors or enhancing the format of the answer. The wrappers integrate code-based rules, APIs, web searches, and small models to efficiently handle editing and customization tasks according to user-defined protocols. \goodname~\customization~ offers several key advantages. It facilitates rapid development and deployment of user-defined protocols, which crucial in production environments where real-time adjustments are necessary. In scenarios where training or fine-tuning LLMs is unfeasible due to time or resource constraints, this method provides an alternative for immediate output customization. Moreover, the wrappers enable flexible incorporation  of various tools and data sources, which enhances the applicability of  \goodname~and reduces resource-intensive LLM calls.

\begin{example}[Warning URLs]\label{example:waring_urls}
The objective was to detect if LLM outputs contain URLs and prepend a warning message of the unsafe URLs at the beginning of the LLM outputs. \customization~should check the safety of the URLs founded,  i.e., whether they are malicious or unreachable, and includes such information in the warning if they were unsafe.
\customization~utilizes a regular expression pattern 
to identify URLs within the text. Upon URLs founded, \customization~calls APIs for detecting phishing URLs, such as Google SafeBrowsing~\citep{google-safe-browsing}, and assess the accessibility of the benign URL by issuing web requests. Malicious URLs, as well as unreachable URLs that return status codes of 4XX, are added in the warning at the beginning of the LLM outputs.
\end{example}

Note that the task in Example~\ref{example:waring_urls} cannot be achieved through prompt engineering when querying LLMs, as the warning must appear at the beginning, and LLMs generate content token by token, making later content unpredictable.
We use the following example to illustrate this property, and experimentally evaluate the efficiency of \customization~wrappers in \textbf{Exp 4} in~\S\ref{sec: exp}.

\begin{example}
\label{ex:llm-limitation}
We present a concrete demonstration of tasks that cannot be reliably accomplished through prompt engineering alone, due to the token-by-token generation mechanism inherent in LLMs. This sequential generation process fundamentally precludes anticipatory knowledge of future token occurrences during text production.
Consider the following prompt submitted to GPT-4:

\begingroup
{
\begin{quote}
\ttfamily
\noindent Write an English poem about a rabbit; please include information at the beginning of the poem about how many times the word ``rabbit'' appears in the poem.
\end{quote}
}
\endgroup

The generated response (shown below) claims four occurrences of ``rabbit,'' while actual analysis reveals five occurrences:
\begingroup
{ 
\begin{quote}
\ttfamily
(The word ``rabbit'' appears 4 times in this poem.) 

In a field where soft grasses sway and dip, 
a \textcolor{red}{rabbit} hopped beneath the moon's silver whip. 
With ears like twin sails in the cool night air, 
each twitch and turn done with utmost care. \\

This \textcolor{red}{rabbit}, cloaked in twilight's gentle hug, 
nestled in shadows, a cozy, earthen snug. 
Dreams of dandelion fields in sunny sprawl, 
where no shadows loom, nor fears enthrall. \\

With a nose that twitches at secrets untold, 
and paws that dance over leaves of gold, 
the \textcolor{red}{rabbit} finds joy in the simplest of things—the warmth of the sun, the freedom of springs. \\

But as dawn breaks and stars bid goodbye, 
back to the burrow, this \textcolor{red}{rabbit} must fly. 
For the world is wide and wolves roam free,
yet in his burrow, safe as can be, this \textcolor{red}{rabbit} shall be.
\end{quote}
}
\endgroup

This discrepancy arises from the LLM's inability to maintain an accurate token count during generation while simultaneously composing coherent text. The example underscores the necessity of post-inference wrappers to detect and correct such errors in model outputs.
\end{example}

\section{\goodname~\fixing}\label{sec:fixing}

\goodname~\fixing~addresses errors in the LLM  outputs that are challenging to resolve through editing with wrappers in~\customization, particularly, hallucinated content. \fixing~analyzes and corrects the hallucinated output based on the reason for the hallucinations generated by the hallucination detection model.

\goodname~\fixing~takes several key inputs, including the user's original query, the context retrieved with \grounding, the hallucinated responses generated by the LLM, as well as the reason for hallucination. 
Given these inputs, \fixing~corrects the flawed output according to the hallucination reason.
To enable \fixing~to handle hallucinations effectively, we  leverage the same hallucination detection dataset as \detection, i.e., HaluEval~\citep{li2023halueval}, that contains user questions, contexts, hallucinated LLM answers, and correct answers. 
We also designed a customized data template that incorporates the information. The data templates for training, inference, as well as an example for the training data, are demonstrated in Figure~\ref{fig: training_data_example_fixing}.

\section{Experiments}\label{sec: exp}

\begin{table*}[htbp]
\small
\caption{Dataset Overview}
\label{tab:exp_datasets}
\begin{center}

\begin{tabular}{cccccp{8.4cm}}
\toprule
\textbf{Dataset} & \textbf{Data size} & \textbf{Train} & \textbf{Validation} & \textbf{Test} & \textbf{Description} \\
\midrule

\multicolumn{6}{c}{\cellcolor[HTML]{EFEFEF}\textbf{\detection~Training Data}} \\ 

HEx-PHI~\citep{anonymous2024finetuning} & 330 & 330 & 0 & 0 & Harmful instructions of 11 prohibited categories.\\

OpenAI Moderation~\citep{openai-data-paper} & 1680 & 160 & 1,500 & 0 & Prompts annotated with OpenAI taxonomy. \\

Hotpot QA~\citep{yang2018hotpotqa} & 113k & 3,000 & 2,500 & 500 & QA pairs based on Wikipedia knowledge. \\

Truthful QA~\citep{lin2021truthfulqa} & 827 & 500 & 100 & 100 & Questions spanning 38 categories, including health, law, politics, etc. \\

Awesome GPT Prompts~\citep{awesome-chatgpt-prompts} & 153 & 0 & 150 & 0 & Awesome prompt examples to be used with ChatGPT. \\

Jigsaw Unintended-Bias~\citep{jigsaw-unintended-bias-in-toxicity-classification} & 2M & 100,000 & 2,000 & 300 & Comment data that contains labels for unsafe content. \\

GPT-Jailbreak~\citep{ChatGPT-Jailbreak-Prompts} & 79 & 0 & 78 & 0 & ChatGPT jailbreak prompts. \\

Jailbreak~\citep{jailbreak-classification} & 1.3k & 400 & 0 & 70 & A dataset that contains jailbreak prompts and benign prompts. \\

Personalization Prompt~\citep{filtered_personalization_prompt_response} & 10.4k & 1,000 & 800 & 200 & Prompt-response pairs for personalized interactions with LLMs. \\

QA-Chat Prompts~\citep{qa-chat-prompts} & 200 & 0 & 200 & 0 & A QA dataset. \\

ChatGPT Prompts~\citep{ChatGPT-prompts} & 360 & 350 & 0 & 0 & Human prompts and ChatGPT responses. \\

10k-Prompts Ranked~\citep{10k_prompts_ranked} & 10.3k & 500 & 500 & 200 & Prompts with quality rankings created by 314 members of the open-source ML community using Argilla, an open-source tool to label data. \\

Iterative Prompt~\citep{iterative-prompt-v1-iter1-20K} & 20k & 500 & 500 & 200 & A dataset of user prompts. \\

Instruction Following~\citep{instruction-following} & 514 & 200 & 340 & 0 & An instruction dataset. \\

\multicolumn{6}{c}{\cellcolor[HTML]{EFEFEF}\textbf{\detection~Evaluation Data}} \\ 

ToxicChat~\citep{lin2023toxicchat} & 10.2k
& - & - & - &  Unsafe content dataset for evaluation. \\

\multicolumn{6}{c}{\cellcolor[HTML]{EFEFEF}\textbf{\grounding~Evaluation Data}} \\ 

E-Commerce~\citep{e_commerce_dataset} & 65 &-&- & -& Use the ``faq'' subset that contains QA pairs between users and agents. \\

PatientDoctor~\citep{patient-doctor-chat-data} & 379k&-&- & - & Dialogue data between doctors and patients. \\

ChatDoctor dataset~\citep{ChatDoctor-dataset} & 119.4k&-&- & - & Dialogue data between doctors and patients. \\

\multicolumn{6}{c}{\cellcolor[HTML]{EFEFEF}\textbf{\fixing~Wrapper Evaluation Data}} \\

E-Commerce~\citep{e_commerce_dataset} & 1.89k &-&- & -& Use the ``faq'' and the ``product'' subsets that contains product descriptions. \\

RedditSYACURL Dataset~\citep{reddit-url-data} & 8.61k &-&- & -& A dataset that contains titles, summaries, and links of articles. \\

\multicolumn{6}{c}{\cellcolor[HTML]{EFEFEF}\textbf{\detection~Hallucination Detection  Data}} \\ 
HaluEval-qa~\citep{li2023halueval} & 10k &8,000&1,500 & 500& 
\multirow{4}{8.6cm}{A hallucination dataset with 3 subsets: 1) ``qa'' with question, right answer, hallucinated answer, and knowledge; 2) ``dialogue'' with dialogue history, right response, hallucinated response, and knowledge; and
3) ``summarization'' with document, right summary, and hallucinated summary.}\\
HaluEval-dialogue~\citep{li2023halueval} & 10k &8,000&1,500 & 500& \\
HaluEval-summarization~\citep{li2023halueval} & 10k &8,000&1,500 & 500& \\
\\

\multicolumn{6}{c}{\cellcolor[HTML]{EFEFEF}\textbf{\fixing~Hallucination Correction Data}} \\ 
HaluEval-qa~\citep{li2023halueval} & 10k &8,000&1,000 & 1,000& 
\multirow{3}{8.6cm}{
The hallucination correction dataset was augmented with a \texttt{hallucination\char`_reason} column, derived from the detection results of \detection. 
}
\\
HaluEval-dialogue~\citep{li2023halueval} & 10k &8,000&1,000 & 1,000& \\
HaluEval-summarization~\citep{li2023halueval} & 10k &8,000&1,000 & 1,000& \\

\bottomrule
\end{tabular}
\end{center}
\end{table*}

We evaluate the performance of different modules in \goodname. 
We use our self-developed model, Fox-1~\cite{fox}, as our base model for finetuning three models, including an unsafe content detection model for \detection, an explainable hallucination detection model for \detection, and a hallucination fixing model for \fixing. 
Below we first introduce Fox-1 and the finetuned the models for different functional modules, then introduce experiment settings, and finally present our evaluation results. 

\noindent\underline{\textit{Base Model. }}
Fox-1 is self-developed, decoder-only transformer-based language model with only 1.6B parameters~\cite{fox}. 
It was trained with a 3-stage data curriculum on 3 trillion tokens of text and code data in 8K sequence length. The base model uses grouped query attention (GQA) with 4 KV heads and 16 attention heads and has a deeper architecture than other SLMs. Specifically, it has 32 transformer decoder blocks, 78\% deeper than Gemma-2B~\cite{team2024gemma}, 33\% deeper than Qwen1.5-1.8B~\cite{qwen} and StableLM-2-1.6B~\cite{bellagente2024stable}, and 15\% deeper than OpenELM-1.1B~\cite{mehta2022cvnets,mehtaOpenELMEfficientLanguage2024}.

\noindent\underline{\textit{Model Finetuning. }}
\detection~model is trained with a combined dataset that extract from 15 datasets to simulate real world unsafe content.
Hallucination detection and explanation model and the hallucination fixing model are trained with HaluEval dataset~\citep{li2023halueval}. 
The datasets for training and evaluation are summarized in Table~\ref{tab:exp_datasets}.

\noindent\underline{\textit{Experimental Setting. }}
We utilized datasets that contain important knowledge to evaluate \grounding, where inaccurate retrieval can cause financial losses or harmful medical advice. We selected E-Commerce dataset~\cite{e_commerce_dataset} that contains customer service interactions on an online platform, and two healthcare datasets, PatientDoctor dataset~\cite{patient-doctor-chat-data} and the ChatDoctor dataset~\cite{ChatDoctor-dataset}, which contain QA pairs between doctors and patients. 
We leveraged \textit{callback} to evaluate the effectiveness of the two indexing methods in \grounding. 
\customization~evaluations are conducted with E-Commerce~\cite{e_commerce_dataset} and RedditSYACURL Dataset~\citep{reddit-url-data}.
The information of the datasets is summarized in Table~\ref{tab:exp_datasets}. Evaluations and model training experiments are conducted on a server with 8 NVIDIA
H100 GPUs.

\begin{figure*}[htbp]
    \centering
    \begin{minipage}{0.3\textwidth}
            \centering
    \includegraphics[width=\linewidth]{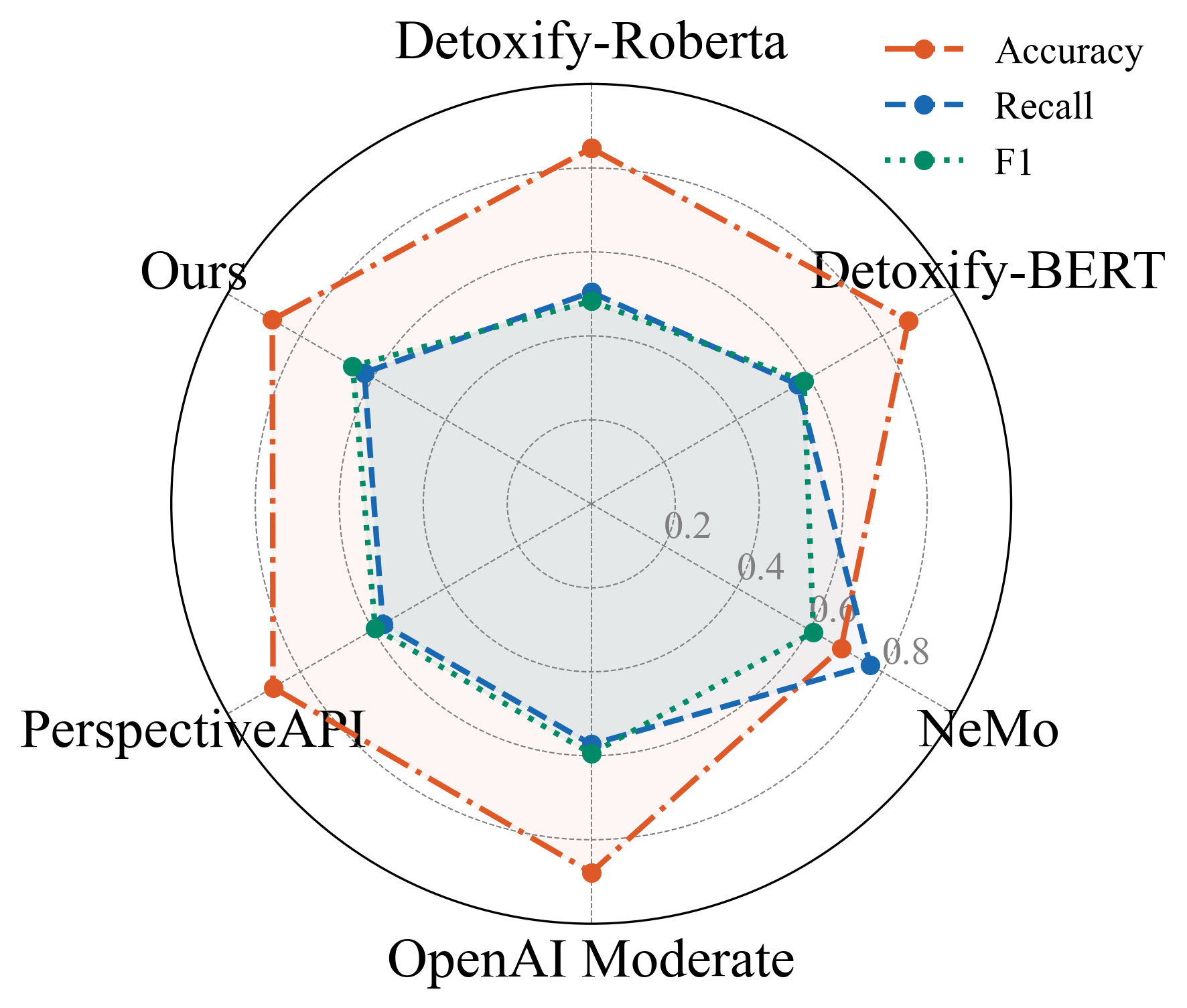}

        \caption{Safety detection}\label{fig:safety_detection_exp}
    \end{minipage}
    \hfill
    \begin{minipage}{0.3\textwidth}
            \centering
    \includegraphics[width=\linewidth]{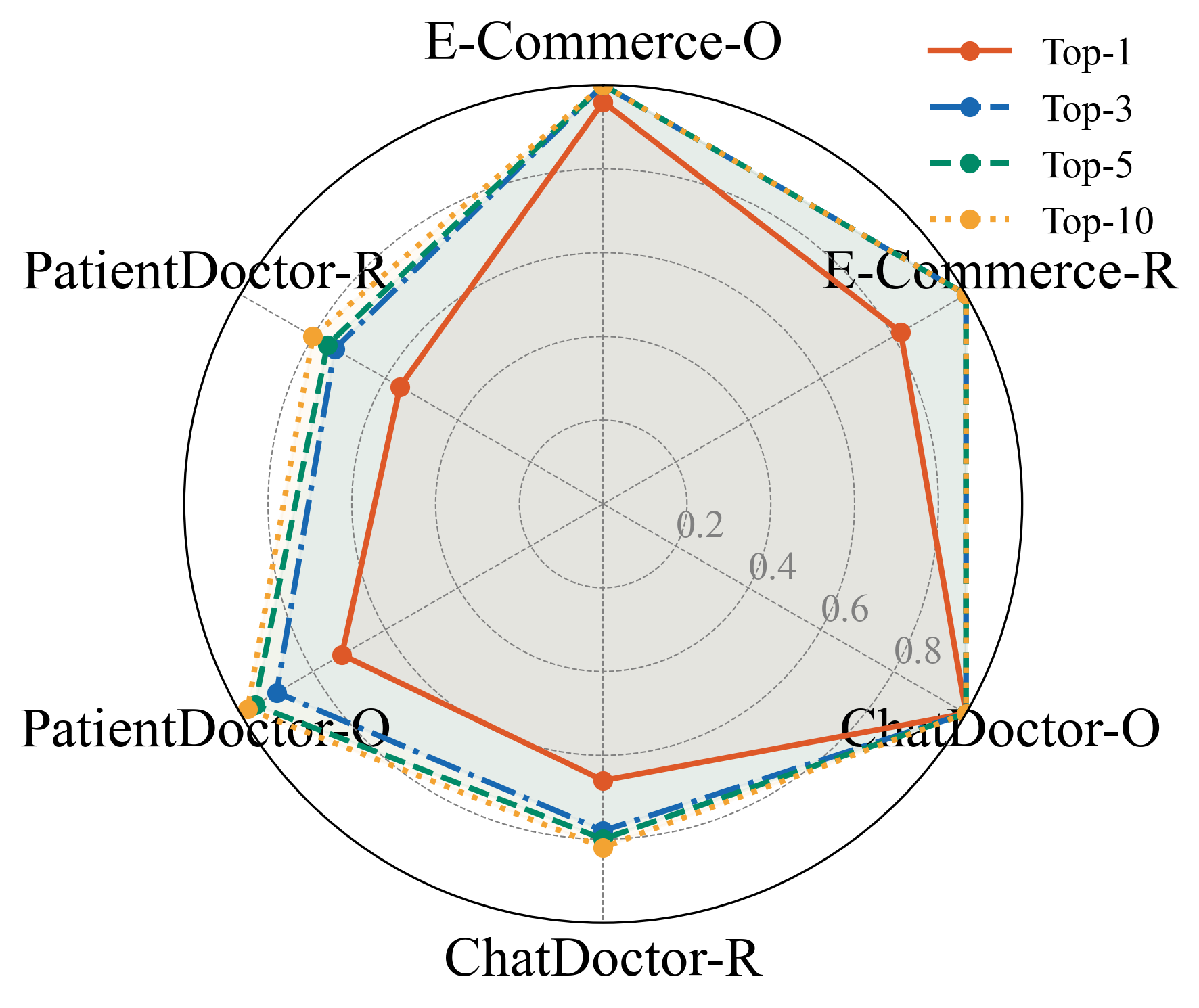}

        \caption{Whole index}\label{fig:whole_knowledge_idx}
    \end{minipage}
    \hfill
    \begin{minipage}{0.3\textwidth}
            \centering \includegraphics[width=\linewidth]{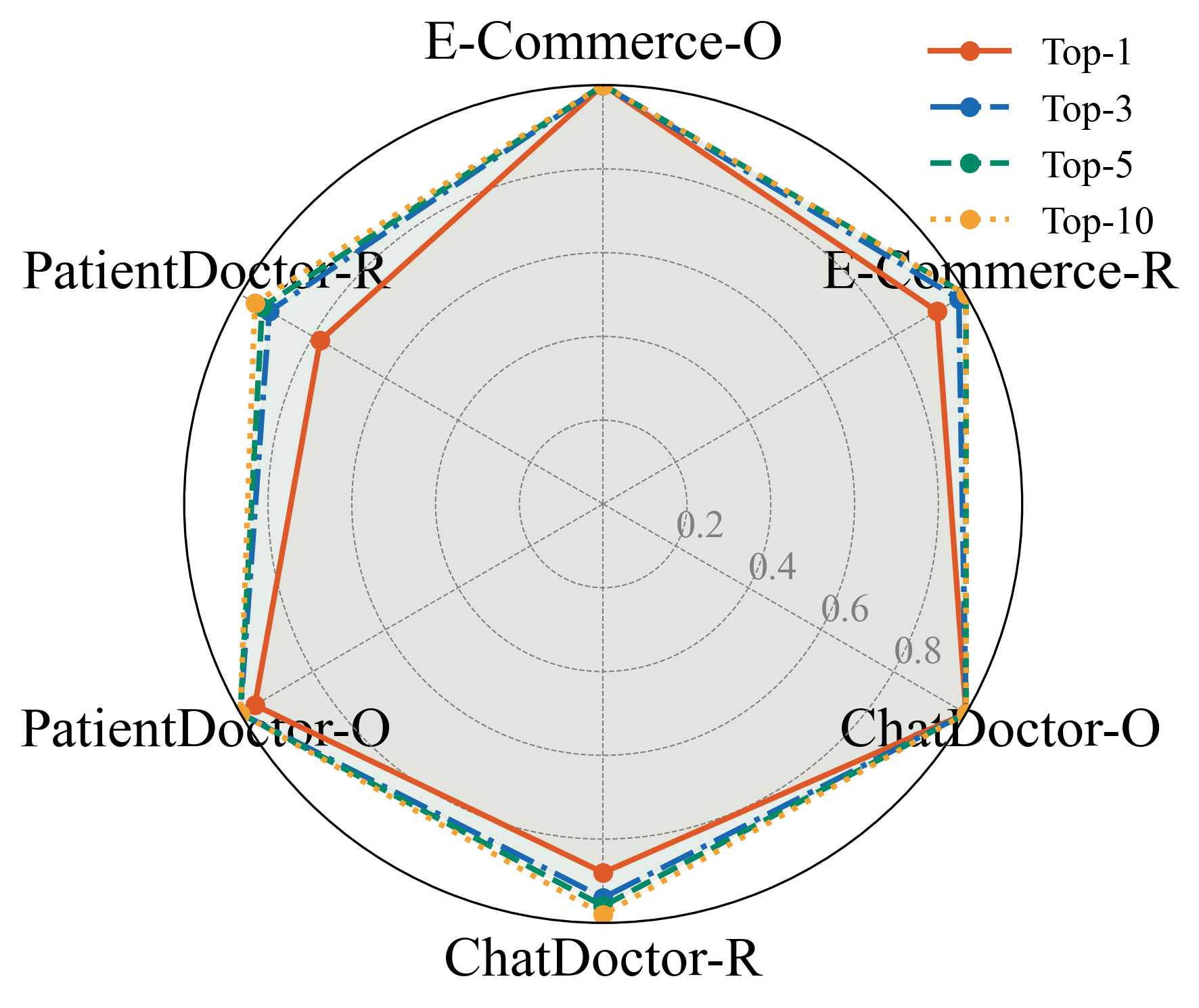}

        \caption{Key index}\label{fig:key_info_idx}
    \end{minipage}

\end{figure*}

\noindent\textbf{Exp 1. Unsafe user inputs detection in~\detection. }
The datasets and the number of records involved in training, validation, and test phases are summarized in Table \ref{tab:exp_datasets}. We compare our approach with 
Detoxify-Roberta~\citep{Detoxify},
Detoxify-BERT~\citep{Detoxify},
Nvidia NeMo GuardRail~\citep{nemo}, 
OpenAI Moderate~\citep{openai-data-paper}, and 
PerspectiveAPI~\citep{perspective-api} in Figure~\ref{fig:safety_detection_exp}.   
Results show that our model achieves comparable performance with OpenAI API. 
Overall, our model demonstrates robust performance across key metrics, indicates its effectiveness and reliability in real-world applications.

\begin{table*}[ht]
\centering
\small
\caption{URL Detection Task}
\label{tab:fixing_exp}
\begin{tabular}{@{}lcccccc@{}}
\toprule
\textbf{Metrics} & \textbf{Ours} & \textbf{TinyLLama} & \textbf{Mistral-7B} & \textbf{LLama2-7B} & \textbf{LLama3-8B} & \textbf{Falcon-40B} \\ \midrule
Avg. Time (s) & \textbf{1.06} & 13.17 & 10.93 & 9.10 & 20.10 & 34.67 \\
Detection Acc. & \textbf{100.00\%} & {\redxmark} {(Fail)} & 91.67\% & 83.33\% & 37.50\% & {\redxmark} {(Fail)} \\
Validation Acc. & \textbf{83.33\%} & {\redxmark} {(Fail)}& 45.83\% & 54.17\% & 37.50\% & {\redxmark} {Fail} \\
\bottomrule
\end{tabular}
\end{table*}

\noindent\textbf{Exp 2. Hallucination detection in LLM outputs in~\detection. } 
We fine-tuned our hallucination detection model using the HaluEval dataset \cite{li2023halueval}. We utilize the three subsets in HaluEval, including \textit{i}) the ``qa'' subset that 
contains question, right answer, hallucinated answer, and knowledge,  \textit{ii}) the ``dialogue'' subset that contains dialogue history, right response, hallucinated response, and knowledge, and  \textit{iii})
the ``summarization'' subset that contains document, right summary, and hallucinated summary.
For each subset, we set 8,000 data samples for training, 1,500 for validation, and 500 for testing.  
Our model achieved an accuracy of 0.78 on the testing data.

\noindent\textbf{Exp 3. Evaluation of different indexing methods in~\grounding. }
To comprehensively evaluate retrieval performance and simulate user queries real-world applications, we used two types of queries, including \textit{i}) \textit{original queries} that match original questions in the datasets (``O'' in Figure~\ref{fig:whole_knowledge_idx} and Figure~\ref{fig:key_info_idx}), and \textit{ii}) \textit{rephrased queries} generated with language models (i.e., TinyLlama~\citep{zhang2024tinyllama} or a summarization  model~\citep{summarization-model}) based on the original questions to simulate variability in user questions (``R'' in Figure~\ref{fig:whole_knowledge_idx} and Figure~\ref{fig:key_info_idx}). 
For each evaluation, we randomly selected 50 questions from the dataset to form a question set $Q$, and processed Top-$k$ queries to compute a callback $C_k(Q)$, where $k$ is set to 1, 3, 5, and 10.
We recorded the callbacks for Whole Knowledge Index and Key Information Index in Figure~\ref{fig:whole_knowledge_idx} and Figure~\ref{fig:key_info_idx}, respectively.
The results indicate that Key Information Indexing outperformed Whole Knowledge Indexing, as key information indexes reflects the user queries better. Also, both original queries and rephrased queries achieved high callback rates, which demonstrates the effectiveness of vector retrieval when handling varied user inputs.

\noindent\textbf{Exp 4. Efficiency of wrappers in~\customization. }
We evaluated the efficiency of \customization~in 
with the URL detection and validation task in Example~\ref{example:waring_urls} in \S\ref{sec:customization}. 
We randomly selected 15 records from the each of the E-Commerce dataset~\citep{e_commerce_dataset} and the RedditSYACURL Dataset~\citep{reddit-url-data}, combined each record to construct texts that contained URLs, and set 20\% probability of inserting some malicious URLs into the text. In implementation, we leveraged Regex pattern for detecting URLs, Google SafeBrowsing~\citep{google-safe-browsing} for detecting malicious URLs, and sent HTTP requests to the safe URLs to verify their reachability.
We compared \customization~with several models, including TinyLLama~\citep{zhang2024tinyllama}, Mistral-7B~\citep{jiang2023mistral}, LLama2-7B~\citep{touvron2023llama}, and Falcon-40B~\citep{falcon40b}. 
The results are shown in Table~\ref{tab:fixing_exp}. We record average time to process one query, the success rate of detecting URLs (Detection Acc.), and the accuracy of identifying unsafe URLs (Validation Acc.).
The results show that~\goodname~\customization~takes much less time (1.06s per query) and
significantly outperforms calling the models for editing LLM outputs. Also, TinyLLama and Falcon-40B failed to detect any URLs in the contents. Though Mistral is able to detect URLs with a high accuracy of 91.67\%, the accuracy of identifying unsafe URLs is only 45.83\%.

\noindent\textbf{Exp 5. Effectiveness of fixing hallucinations in~\fixing. } 
We fine-tuned our fixing model using the HaluEval dataset \cite{li2023halueval}. We selected the QA and dialogue subsets. For each subset, we utilized 8,000 data samples for training, 1,000 for validation, and 1000 for testing.  
Also, we augmented the hallucination correction dataset with a \texttt{hallucination\char`_reason} column, derived from the detection results of \detection. Such annotation categorizes root causes of hallucinations identified during the detection phase, enabling mitigation strategies in the fixing stages.
We utilize Vectara hallucination detection model~\citep{vectara_halu} for evaluating the consistency between the LLM outputs and the information provided in the original data, including the user questions, the contexts, and the correct answers. We utilized the 100 records in the test dataset of the HaluEval-QA dataset for evaluation. Results show that our fixing model improves the quality of the LLM outputs by a lot. Moreover, 80.7\% of the hallucinated data were fixed using \fixing.

\section{Conclusion}\label{sec:conclusion}

The increasing development of LLMs demands robust safeguards against safety risks  such as toxicity, hallucinations, and adversarial attacks during LLM inference.  
While existing solutions often address safety risks in isolation, they fail to 
mitigate safety
risks from a global perspective. \goodname~addresses these challenges through a guardrail pipeline that integrates different functional modules for detection, contextualization, correction, and customization. 
It not only bridges the safety gap in current LLM deployments but also sets a foundation for future research in trustworthy AI. 
Potential directions include extending its modular design to emerging threats, optimizing resource efficiency for low-latency applications, and integrating multimodal safety checks.

\bibliographystyle{ACM-Reference-Format}
\bibliography{reference}

\end{document}